\documentclass[runningheads]{llncs}
\usepackage[utf8]{inputenc} 
\usepackage{graphicx}
\usepackage{marvosym}
\usepackage{epstopdf}
\usepackage{amsfonts}
\usepackage{amsmath}

\begin{document}
\title{Characterization Multimodal Connectivity of Brain Network by Hypergraph GAN  for Alzheimer's Disease Analysis}

\author{Junren Pan\inst{1} \and
Baiying Lei\inst{2} \and
Yanyan Shen\inst{1} \and
Yong Liu\inst{3} \and
Zhiguang Feng\inst{4} \and
Shuqiang Wang\inst{1}\textsuperscript{(\Letter)}
}
\authorrunning{J. Pan et al.}
%
\institute{Shenzhen Institutes of Advanced Technology, Chinese Academy of Sciences, Shenzhen
518000, China\\
\email{\{jr.pan,yy.shen,sq.wang\}@siat.ac.cn}  \\
\and Shenzhen University, Shenzhen, 518000, China\\
\email{leiby@szu.edu.cn}   \\
\and Renmin University of China, Beijing, 100000, China  \\
\email{liuyonggsai@ruc.edu.cn}
\and Harbin Engineering University, Haerbin, 150000 , China\\
\email{fengzhiguang@hrbeu.edu.cn}
}
\titlerunning{Hypergraph GAN}

%
%
%
%
%
\maketitle              
\begin{abstract}
Using multimodal neuroimaging data to characterize brain network is currently an advanced technique for Alzheimer's disease(AD) Analysis.
Over recent years the neuroimaging community has made tremendous progress in the study of resting-state functional magnetic resonance imaging (rs-fMRI)
derived from blood-oxygen-level-dependent (BOLD) signals and Diffusion Tensor Imaging (DTI) derived from white matter fiber tractography.
However, Due to the heterogeneity and complexity between BOLD signals and fiber tractography, Most existing multimodal data fusion algorithms can not sufficiently take advantage of the complementary information between rs-fMRI and DTI.
To overcome this problem, a novel Hypergraph Generative Adversarial Networks(HGGAN) is proposed in this paper, which utilizes Interactive Hyperedge Neurons module
(IHEN) and Optimal Hypergraph Homomorphism algorithm(OHGH) to generate multimodal connectivity of Brain Network from rs-fMRI combination with DTI.
To evaluate the performance of this model, We use publicly available data from the ADNI database to demonstrate that the proposed model not only can identify discriminative brain regions of AD but also can effectively improve classification performance.

\end{abstract}
\section{Introduction}

Alzheimer's disease (AD) is an irreversible, chronic neurodegenerative disease, and is the main reason for dementia among aged people~\cite{DadarPascoal2017}.
Those people who suffered from AD will gradually lose cognitive function such as remembering or thinking, and eventually lose the ability to perform daily activates~\cite{Association2018,a2}.
During the past few decades, there are two methods that are widely applied to diagnose AD or other neurodegenerative diseases. One is resting-state functional magnetic resonance imaging(rs-fMRI)~\cite{HuettelSong2004},which is based on blood-oxygen-level-dependent(BOLD)signals; the other is Diffusion Tensor Imaging(DTI)~\cite{Westlye2010}, which uses the white matter fiber tractography.
But unfortunately, the cause and mechanism of AD are still not entirely clear.
To overcome such barriers, new tools should be implemented. One of the modern approaches is the analysis of the brain network.
Brain network is a representation of connectivity maps between brain regions(defined in an anatomical parcellation, or brain atlas), which can give essential insights in studying brain science.
Specifically, Brain network provides enormous information about global and local features of neuronal network architecture, which can evaluate the disease status, identify crucial brain regions of AD, and reveal the
mechanism of AD.
In terms of mathematics, a brain network is equivalent to a weighted undirected graph, where each node in the graph corresponds to a brain region,
and the value of edge between two different nodes represents the strength of connectivity of the corresponding brain regions.
As the development of deep learning technology for neuroimaging data provides powerful tools to compute and analyze the brain network, many studies\cite{sh3,c1,c2,c3,JeonKang2020,a1,ktz5,ktz6,ktz7,ktz12,ktz13} have exploited deep learning models to
obtain AD-related features from brain network.
Recent studies\cite{ZhangShen2012,sh2,LeiCheng2020,ktz8,ktz9,ktz10,ktz11} have shown that the combination of multimodal neuroimging data can discover complementary information of brain network, which is beneficial to improve the deep feature representation.
Therefore, designing an effective multimodal fusion model for computing brain network has become a hot topic.
However, most existing multimodal fusion models directly use neural fiber tractography (i.e., structural connectivity, short for SC) to determine the edges between brain regions and use the signal of brain activities (i.e., BOLD) to characterize the nodes feature.
Due to the heterogeneity and complexity between BOLD signals and fiber tractography, such methods can not sufficiently take advantage of the complementary information.
Moreover, the previous studies have demonstrated that brain cognitive mechanisms involve multiple co-activated brain regions (i.e., neural circuit) interactions rather than single pairwise interactions\cite{LeeSmyser2013,Cao2018}. To overcome the above problem, some researchers began to use hypergraph to characterize brain network~\cite{Munsell2016}. Hypergraphs are a generalization of graphs, which are increasingly important in data science thanks to the development of combinatorial mathematics and computer science. The most significant difference between hypergraphs and graphs is that each hyperedge in a hypergraph can contain multiple nodes.
This feature of hypergraphs is a natural superiority to analyze the neural circuit in brain network.
On the other hand, Generative Adversarial Network (GAN)~\cite{GAN,g1,ktz1,ktz2,ktz3,ktz4} is nowadays the dominant method for processing multimodal medical images. GAN has a strong ability to deal with unsupervised learning tasks.
Motivated by this advantage, we devise a hypergraph Generative Adversarial Networks (HGGAN) to characterize the multimodal connectivity of brain network based on the hypergraph theory for AD analysis.

In this paper, we propose a HGGAN to generate multimodal connectivity from rs-fMRI and DTI. We design the generator of HGGAN using the Interactive Hyperedge Neurons(IHEN) module, which is an improvement of recent work HNHN~\cite{Bengio}. Compared with classical CNN-based GAN that can only operate on Euclidean data, our proposed model can learn intrinsic relationships and complementary information based on hypergraph structure multimodal imaging data. 

\section{Method}
\begin{figure}
\includegraphics[width=\textwidth]{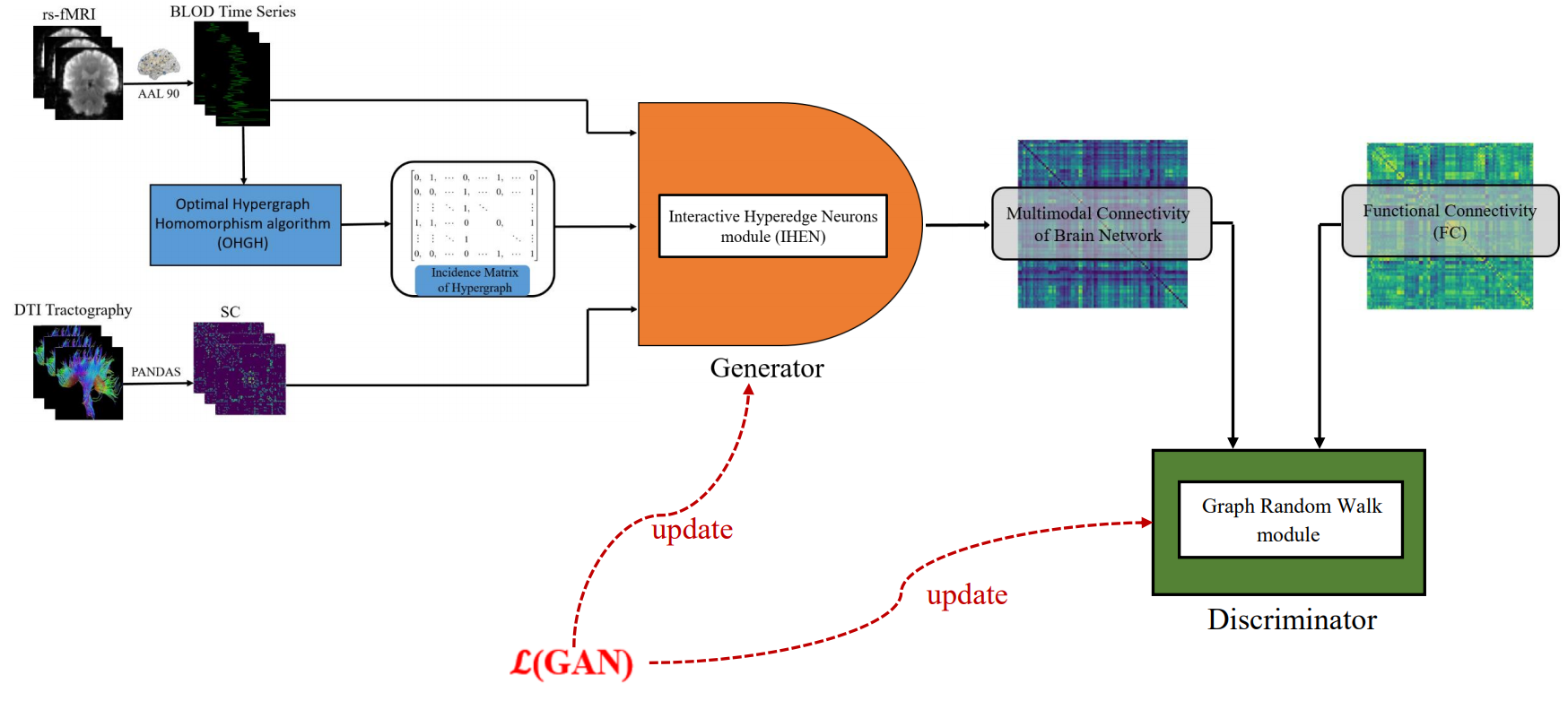}
\caption{The framework of the proposed HGGAN model.
} \label{fig1}
\end{figure}

Figure~\ref{fig1} shows the proposed framework for multimodal connectivity generation. Specifically, we first extract the BOLD time series of each brain region (90 regions
in total) from rs-fMRI data by using AAL atlas~\cite{aal}. We constructed SC from DTI data by using PANDA~\cite{panda}.
We use BOLD to: (i) represent the features of nodes and (ii) calculate the incidence matrix of the hypergraph through the OHGH algorithm.
And we use SC to calculate the features of hyperedges. In this way the features of nodes, the incidence matrix, and the features of hyperedges were fed into the generator to generate the multimodal connectivity.
Simultaneously, we utilize GRETNA~\cite{gretna} to obtain functional connectivity (FC), and FC is used as real samples to train the discriminator.
\subsection{Data and Pre-processing}
We select a total of 219 subjects from the ADNI database.
The composition of our dataset is given as follows:
Our dataset includes 18 AD female patients and 32 AD male patients;
13 late mild cognitive impairment (LMCI) female patients and 12 LMCI male patients;
24 early MCI (EMCI) female patients and 45 EMCI male patients;
43 normal control (NC) female and 32 male NC.
The average age of AD LMCI, EMCI and NC is 75.3, 74.9, 75.8 and 74.0, respectively.

Preprocessing of rs-fMRI data uses the DPARSF~\cite{dparsf} toolbox and the GRETNA toolbox based on the data analysis software \emph{statistical parameter mapping (SPM12)}~\cite{spm}.
First we use DPARSF to preprocess the initial DICOM format of rs-fMRI data into NIFTI format data. We apply the standard steps for rs-fMRI data preprocessing, including the discarding of the first 20 volumes, head motion correction, spatial normalization,  and Gaussian smoothing in this stage.
Then we use the AAL atlas to divide brain space into 90 brain regions-of-interest (ROIs).
Finally, we send these NIFTI format data into the brain network analysis toolbox GRETNA to extract the BOLD time series corresponding to  ROIs.
We utilized DPARSF to obtain BOLD time series and we utilized GRETNA to obtain FC matrix.
Also by using the AAL atlas,
Preprocessing of DTI data uses the PANDA toolbox so that the number of white matter fiber tractography can be regarded as strength of physical
connections in the $90\times90$ SC matrix.




\subsection{Hypergraph and Optimal Hypergraph Homomorphism Algorithm}
We first introduce the basic concepts and notations of hypergraphs. We define a hypergraph $H=(V,E)$ to consist of a set $V$ of nodes and set $E$ of hyperedges,
where each hyperedge itself is a set of nodes. Let $n = |V|$  and $m = |E|$. We label the nodes as $v_i$ for $i\in \{1,\cdots,n\}$, and the hyperedges as $e_j$ for $j\in\{1,\cdots,m\}$.
The incidence matrix $A\in\mathbb{R}^{n\times m}$ of hypergraph $H$ is denoted by
\[
A(i,j)=
\begin{cases}
1& v_i\in e_j\\
0& v_i\notin e_j
\end{cases}
\]

\begin{figure}
\centering
\includegraphics[width=0.7\textwidth]{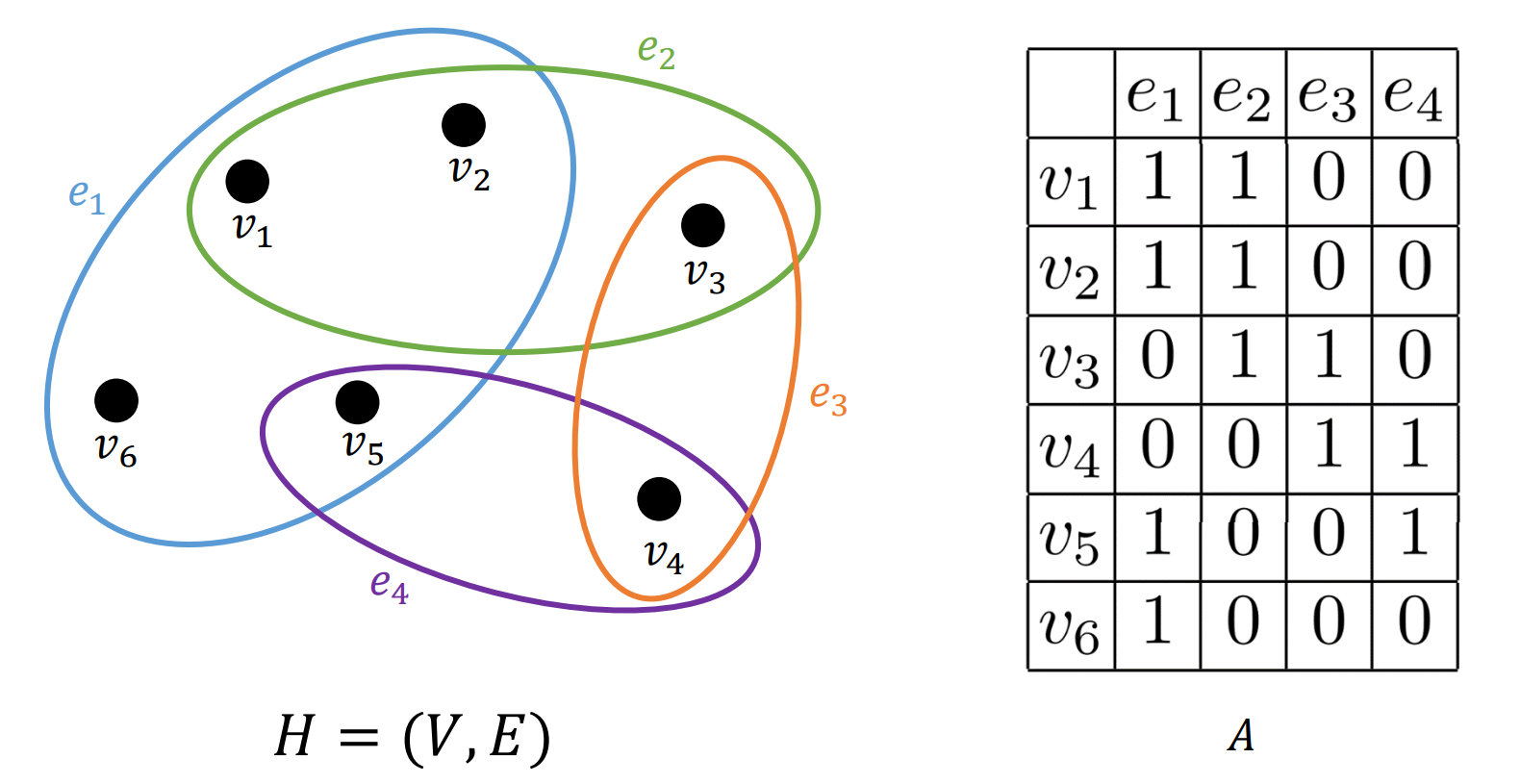}
\caption{Left: a hypergraph with $V=(v_1,v_2,v_3,v_4,v_5.v_6)$ and $E={e_1,e_2,e_3,e_4}$, where $e_1=\{v_1,v_2,v_5,v_6\}$, $e_2=\{v_1,v_2,v_3\}$, $e_3=\{v_3,v_4\}$, $e_3=\{v_4,v_5\}$. Right: its incidence matrix.}
\end{figure}

In this paper, we regard each ROI as a node. We apply the Dynamic Hypergraph Construction algorithm (DHC)~\cite{DynamicHG} to calculate the initial hypergraphs $\{H_k'\}_{k=1}^{219}$ for every subject.
However, each initial hypergraph $H_k'$ only focuses on representing multiple co-activated ROIs information for the corresponding subject.
Therefore, if we directly send initial hypergraphs $\{H_k'\}$ to the generator to obtain multimodal connectivity, the robustness of the generation result cannot be ensured.
To overcome this shortage, we proposed a novel Optimal Hypergraph Homomorphism algorithm(OHGH). OHGH can calculate the hypergraph $H$ which have optimal homomorphism with respect to the initial hypergraphs $\{H_k'\}$.
Before introducing OHGH, we first define the similarity on hypergraphs. Given two hypergraphs $H=(V,E)$ and $H'=(V,E')$ with the same nodes and $|E|=|E'|$, the similarity between $H, H'$ is defined by
\[
\operatorname{Sim}(H,H'):=\sup_{\substack{f:E\rightarrow E'\\f\;  \text{is surjective} }}\frac{1}{|E|}\sum_{e\in E}s(e,f(e))
\]
where
\[
s(e,e') = \frac{|\mathcal{V}_{H}(e)\cap \mathcal{V}_{H'}(e')|}{|\mathcal{V}_{H}(e)\cup \mathcal{V}_{H'}(e')|},
\]
\[
\mathcal{V}_{H}(e) = \{v| v\in e \}, \quad \mathcal{V}_{H'}(e') = \{v| v\in e' \}.
\]
Now we can give OHGH as follows
\[
H = \operatornamewithlimits{argmax}_{H}\sum_{k} \operatorname{Sim}(H,H_k')
\]
With the optimal homomorphism hypergraph $H$, we construct the hypergraphs $\{H_k\}$ for every subject by concatenating $H$ and $H_k'$ (i.e., $H_k = H||H_k'$).
Furthermore, we obtain the incidence matrices $\{A_k\}$ based on $\{H_k\}$.

\subsection{Generator and Interactive Hyperedge Neurons Module}
For each subject, say for the $k$-th subject, we use the BOLD time series $B_k \in \mathbb{R}^{90\times d}$ of the $k$-th subject to represent the nodes features and we rewrite it as $X_{k,V}^{(0)}$ (where $90$ is the number of nodes, $d$ is the length of the BOLD time series).
Meanwhile, we calculate the hyperedges features $X_{k,E}^{(0)}$ by using the SC matrix $S_k\in \mathbb{R}^{90\times 90}$ and the incidence matrix $A_k \in \mathbb{R}^{90\times m}$ of the $k$-th subject (where $m$ is the number of hyperedges). Specifically, it is computed in the following way
\[
X_{k,E}^{(0)} = A_k^T  S_k
\]
The proposed generator $G$ is composed of $L$ multi-layer Interactive Hyperedge Neurons modules (IHEN).
The definition of IHEN is given as follows
\[
X_{k,V}^{(l+1)} = \sigma(A_k X_{k,E}^{(l)} W_{E}^{(l)}+ \lambda X_{k,V}^{(l)}W_{V}^{(l)}),
\]
\[
X_{k,E}^{(l+1)} = \sigma(A_k^{T} X_{k,V}^{(l)} W_{V}^{(l)}+ \lambda X_{k,E}^{(l)}W_{E}^{(l)}),
\]
where $X_{k,V}^{(l)}$ and $X_{k,E}^{(l)}$ are the $l$-th layer features of nodes and hyperedges, $W_{V}^{(l)}$  and $W_{E}^{(l)}$ are the $l$-th layer weight matrices of nodes and hyperedges, $\lambda$ is hyperparameters,  $l = 0,2,\cdots, L-1.$

We compute the hyperedges weights and the hyperedge-independent nodes weights by using $X_{k,V}^{(L)}$ and $X_{k,E}^{(L)}$ as follows:
\[
\gamma_{k,e_j}(v_i) = A_k(i,j)\langle X_{k,V}^{(L)}(i,\cdot),  X_{k,E}^{(L)}(j,\cdot) \rangle
\]
\[
w_{k}(e_j) = \|X_{k,E}^{(L)}(j,\cdot)\|_2
\]
where $\gamma_{k,e_j}(v_i)$ is the weight of the $i$-th node with respect to the $j$-th hyperedge for the $k$-th subject and $w_{k}(e_j)$ is the weight of the the $j$-th hyperedge for the $k$-th subject. From this, we obtain the multimodal connectivity matrix $M_k$ for the $k$-th subject by the formulas
\begin{equation}\label{1}
M_k(i_1,i_2) = \sum_{j=1}^{m} \gamma_{k,e_j}(v_{i_1})w_k(e_j)\gamma_{k,e_j}(v_{i_2})
\end{equation}

Combining the multimodal connectivity matrix $M_k$ with the nodes features $X_{k,V}^{(L)}$ , we give the nodes correlation coefficients as follows:
\begin{equation}\label{2}
\text{Co}_{k}(v_i) = \frac{1}{90}\sum_{j=1}^{90} M_k(i,j)\langle X_{k,V}^{(L)}(i,\cdot),  X_{k,V}^{(L)}(j,\cdot) \rangle.
\end{equation}

\subsection{Discriminator and Loss Function}
The discriminator $D$ is a normal MLP  architecture, which can compare the difference of distributions between the FC matrix and the multimodal connectivity matrix by using random walk. Fixing an arbitrary node $v_{r_0}$, we take $v_{r_0}$ as the starting point for the random walk. The termination condition of the random walk is that the point comes back to the position of the previous step at some step of random walk. For example, a random walk having $T$-steps and eventually stopping at the $v_{r_T}$ means that the path $[v_{r_0},v_{r_1},\cdots, v_{r_{T-1}},v_{r_{T}},v_{r_{T+1}}]$ satisfies $v_{r_{T-1}} = v_{r_{T+1}}$, and we say the $v_{r_{T}}$ is the endpoint of the path. For the FC matrix of the $k$-th subject, the probability of walking through the path $[v_{r_0},v_{r_1},\cdots, v_{r_{T-1}},v_{r_{T}},v_{r_{T+1}}]$ is given as follows:
\[
p_{real,k}([v_{r_0},v_{r_1},\cdots, v_{r_{T-1}},v_{r_{T}},v_{r_{T+1}}]) = \Big(\prod_{t=1}^{T} \frac{\text{FC}_{k}(r_{t-1},r_{t})}{\|\text{FC}_{k}(r_{t-1},\cdot)\|_1}\Big)\cdot \frac{\text{FC}_{k}(r_{T},r_{T-1})}{\|\text{FC}_{k}(r_{T},\cdot)\|_1}.
\]
Let $\text{Path}(v_{r_0},v)$ be the set of paths with the starting point at $v_{r_0}$ and the endpoint at $v$, then the conditional probability about $v_{r_0}$ is given as follows
\[
p_{real,k}(v|v_{r_0}) = \sum_{\text{path}\in \text{Path}(v_{r_0},v)} p_{real,k}(\text{path})
\]
Similarly, for the multimodal connectivity matrix of the $k$-th subject, the probability of walking through the path $[v_{r_0},v_{r_1},\cdots, v_{r_{T-1}},v_{r_{T}},v_{r_{T+1}}]$ is given by
\[
p_{G,k}([v_{r_0},v_{r_1},\cdots, v_{r_{T-1}},v_{r_{T}},v_{r_{T+1}}]) = \Big(\prod_{t=1}^{T} \frac{M_{k}(r_{t-1},r_{t})}{\|M_{k}(r_{t-1},\cdot)\|_1}\Big)\cdot \frac{M_{k}(r_{T},r_{T-1})}{\|M_{k}(r_{T},\cdot)\|_1}
\]
and conditional probability about $v_{r_0}$ is that
\[
p_{G,k}(v|v_{r_0}) = \sum_{\text{path}\in \text{Path}(v_{r_0},v)} p_{G,k}(\text{path})
\]

The loss function of discriminator is formulated by
\[
\max_{D}\sum_{k=1}^{K}\sum_{i=1}^{90}\mathbb{E}_{v\sim p_{real,k}(v|v_i)}[\log D(v,v_i)]+\mathbb{E}_{v\sim p_{G,k}(v|v_i)}[\log (1-D(v,v_i))].
\]
where $K$ represents the total number of subject in the training dataset.

The loss function of the generators depends on the feedback of discriminator, which is formulated by
\[
\max_{G}\sum_{k=1}^{K}\sum_{i=1}^{90}\mathbb{E}_{v\sim p_{G,k}(v|v_i)}[\log D(v,v_i)].
\]

\section{Experiments}
In order to evaluate the performance of the multimodal connectivity matrix(MC) generated by our proposed method, six binary classification experiments including
AD vs. NC, AD vs. EMCI , AD vs. LMCI, EMCI vs. NC , LMCI vs. NC, and EMCI vs. LMCI are designed to validate the prediction performance.
We use prediction accuracy (ACC), sensitivity (SEN), specificity (SPE) to measure the results of our experiments.

\begin{table}
\caption{Prediction performance in AD vs. NC, AD vs. EMCI and AD vs. LMCI.}\label{tab1}
\resizebox{\textwidth}{25mm}{
\begin{tabular}{|l|l|l|l|l|l|l|l|l|l|l|}
	\hline
	\textbf{Classifier} \quad & \textbf{Connectivity Matrix}  \quad & \multicolumn{3}{|c|}{\textbf{AD vs. NC}}  & \multicolumn{3}{|c|}{\textbf{AD vs. EMCI}} & \multicolumn{3}{|c|}{\textbf{AD vs. LMCI}} \\ \cline{3-11}
    \quad & \quad & ACC \; & SEN \; & SPE \;& ACC \;& SEN \;& SPE \;& ACC \;& SEN \;& SPE  \;\\
    \hline
    \hline
    MLP & SC &79.06 &52.94 & \textbf{96.15}&78.04 & \textbf{88.23} & 65.38& 60.00 & \textbf{76.47} & 25.00 \\ \cline{2-11}
    \quad & FC & 74.41 & 52.94 & 88.46 &68.29 & 64.70 & 65.38 & 56.00 & 64.70 & 37.50 \\ \cline{2-11}
    \quad & MC(Ours) & \textbf{81.39} & \textbf{64.70} & 92.30 &\textbf{80.48} & 76.47 & \textbf{76.92}& \textbf{64.00} & 58.82 & \textbf{75.00} \\
    \hline
    SVM & SC &83.72 &70.58 & \textbf{92.30} &80.48 & 64.70 & \textbf{84.61} & 64.00 & 64.70 & 62.50 \\ \cline{2-11}
    \quad & FC & 79.06 & 70.58 & 84.61 & 73.17 & 52.94 & 80.76 & 56.00 & 52.94 & 62.50 \\ \cline{2-11}
    \quad & MC(Ours) & \textbf{86.04} & \textbf{82.35} & 88.46 & \textbf{82.92} & \textbf{76.47} & 80.76 & \textbf{68.00} & \textbf{70.58} & 62.50 \\
    \hline
    RF & SC &81.39 &70.58 & \textbf{88.46} & 78.04 & 64.70 & \textbf{80.76} & 60.00 & 70.58 & \textbf{37.50} \\ \cline{2-11}
    \quad & FC & 76.74 & 70.58 & 80.76 & 73.17 & 58.82 & 76.92 &60.00 & 70.58 & \textbf{37.50} \\ \cline{2-11}
    \quad & MC(Ours) & \textbf{86.04} & \textbf{88.23} & 84.61 & \textbf{82.92} & \textbf{82.35} & 76.92 & \textbf{64.00} & \textbf{88.23} & 12.50 \\
    \hline
    GCN & SC &86.04 &94.11 & 80.76 & 85.36 & \textbf{94.11} & 73.07 & 68.00 & 76.47 & \textbf{50.00} \\ \cline{2-11}
    \quad & FC & 83.72 & 94.11 & 76.92 &80.48 & 70.58 & 80.76 & 64.00 & 76.47 & 37.50 \\ \cline{2-11}
    \quad & MC(Ours) & \textbf{93.02} & 94.11 & \textbf{92.30} & \textbf{90.24} & 82.35 & \textbf{88.46} & \textbf{72.00} & \textbf{82.35} & \textbf{50.00} \\
    \hline
\end{tabular}}
\end{table}

\begin{table}
\caption{Prediction performance in EMCI vs. NC , LMCI vs. NC and EMCI vs. LMCI.}\label{tab2}
\resizebox{\textwidth}{23mm}{
\begin{tabular}{|l|l|l|l|l|l|l|l|l|l|l|}
	\hline
	\textbf{Classifier} \quad & \textbf{Connectivity Matrix}  \quad & \multicolumn{3}{|c|}{\textbf{EMCI vs. NC}}  & \multicolumn{3}{|c|}{\textbf{LMCI vs. NC}} & \multicolumn{3}{|c|}{\textbf{EMCI vs. LMCI}} \\ \cline{3-11}
    \quad & \quad & ACC \; & SEN \; & SPE \;& ACC \;& SEN \;& SPE \;& ACC \;& SEN \;& SPE  \;\\
    \hline
    \hline
    MLP & SC & 66.00 & 62.50 & \textbf{69.23} & 76.47 & 50.00 & \textbf{84.61} &  68.75 & \textbf{79.16} & 37.50  \\ \cline{2-11}
    \quad & FC & 64.00 & 70.83 & 57.69& 73.52 & 75.00 & 73.07 &  71.87 & 75.00 & \textbf{62.50} \\ \cline{2-11}
    \quad & MC(Ours) & \textbf{70.00} & \textbf{75.00} & 65.38 & \textbf{79.41} & \textbf{87.50} & 76.92 &  \textbf{75.00} & \textbf{79.16} & \textbf{62.50} \\
    \hline
    SVM & SC & 68.00 & 66.66 & \textbf{69.23} & 82.35 & 62.50 & \textbf{88.46} &  71.87 & 83.33 & 37.50 \\ \cline{2-11}
    \quad & FC & 70.00 & 70.83 & \textbf{69.23} & 82.35 & 75.00 & 84.61 &  71.87 & \textbf{87.50} & 25.00 \\ \cline{2-11}
    \quad & MC(Ours) & \textbf{72.00} & \textbf{83.33} & 61.53 & \textbf{85.29} & \textbf{87.50} & 84.61 &  \textbf{78.12} & \textbf{87.50} & \textbf{50.00} \\
    \hline
    RF & SC & 68.00 & 58.33 & 76.92 & 76.47 & 62.50 & 80.76 &  71.87 & 75.00 & 62.50 \\ \cline{2-11}
    \quad & FC & 68.00 & 62.50 & 73.07 & 79.41 & 62.50 & \textbf{84.61} &  75.00 & \textbf{83.33} & 50.00 \\ \cline{2-11}
    \quad & MC(Ours) & \textbf{74.00} & \textbf{66.66} & \textbf{80.76} & \textbf{82.35} & \textbf{87.50} & 80.76 &  \textbf{81.25} & 79.16 & \textbf{87.50} \\
    \hline
    GCN & SC & 72.00 & 70.83 & 73.07 & 85.29 & \textbf{87.50} & 84.61 &  78.12 & 83.33 & 62.50 \\ \cline{2-11}
    \quad & FC & 74.00 & 75.00 & 73.07 & 79.41 & 75.00 & 80.76 &  81.25 & \textbf{91.66} & 50.00 \\ \cline{2-11}
    \quad & MC(Ours) & \textbf{80.00} & \textbf{79.16} & \textbf{80.76} & \textbf{91.17} & \textbf{87.50} & \textbf{92.30} &  \textbf{90.62} & \textbf{91.66} & \textbf{87.50} \\
    \hline
\end{tabular}}
\end{table}


Table~\ref{tab1} and Table~\ref{tab2} summarize the prediction results in different classifiers by using structural connectivity matrix (SC), functional connectivity matrix (FC), and our proposed multimodal connectivity matrix (MC) in eq.\eqref{1}.
Classifiers include Multiple Layer Perception (MLP)~\cite{mlp}, Support Vector Machine (SVM)~\cite{svm}, Random Forest (RF)~\cite{rf}, and Graph Convolution Networks (GCN)~\cite{gcn}.
In detail, the parameters of MLP set as follows:  3-layers with 16, 16, 2 neurons, ReLU activation, 0.001 learning rate. The parameters of SVM set as follows: Gaussian Kernel, 0.15 kernel coefficient. The parameters of RF set as follows: 400 trees, 3 maximum depth.  The parameters of GCN parameters set as follows: 3 Chebyshev graph convolutional layers, 0.1 dropout rate,  0.0001 learning rate. The dataset is randomly split into 65\% for training and 35\% for testing.
All the models are randomly initialized for 5 times. The result shows that the classification accuracy is significantly improved by using the proposed MC matrix, which means the complementary information of different modal is successfully extracted from the proposed method.



\begin{figure}
\includegraphics[width=1.075\textwidth]{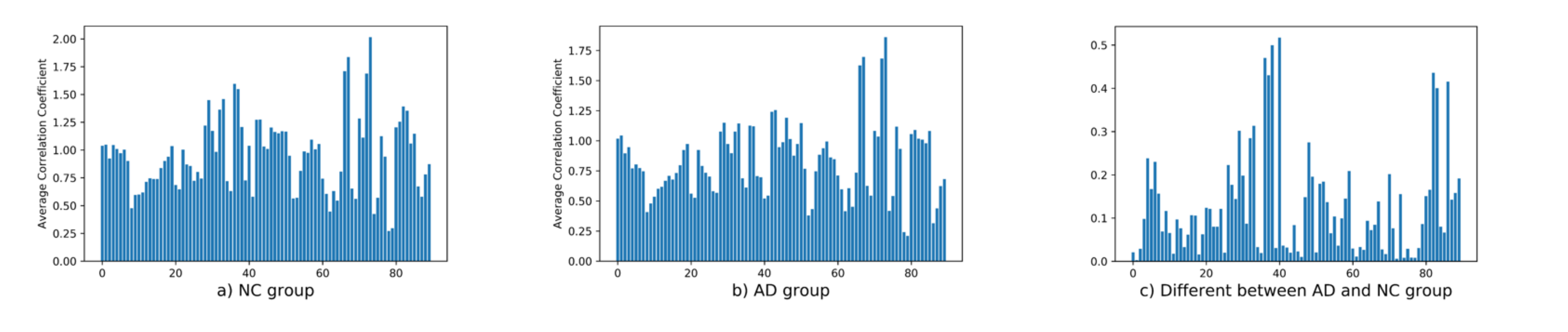}
\caption{In this figure, (a) and (b) shows the average correlation coefficient of each node of  the NC group and AD group. (c) shows the different between the AD group and NC group. } \label{fig2}
\end{figure}


\begin{figure}
\centering
\includegraphics[width=0.85\textwidth]{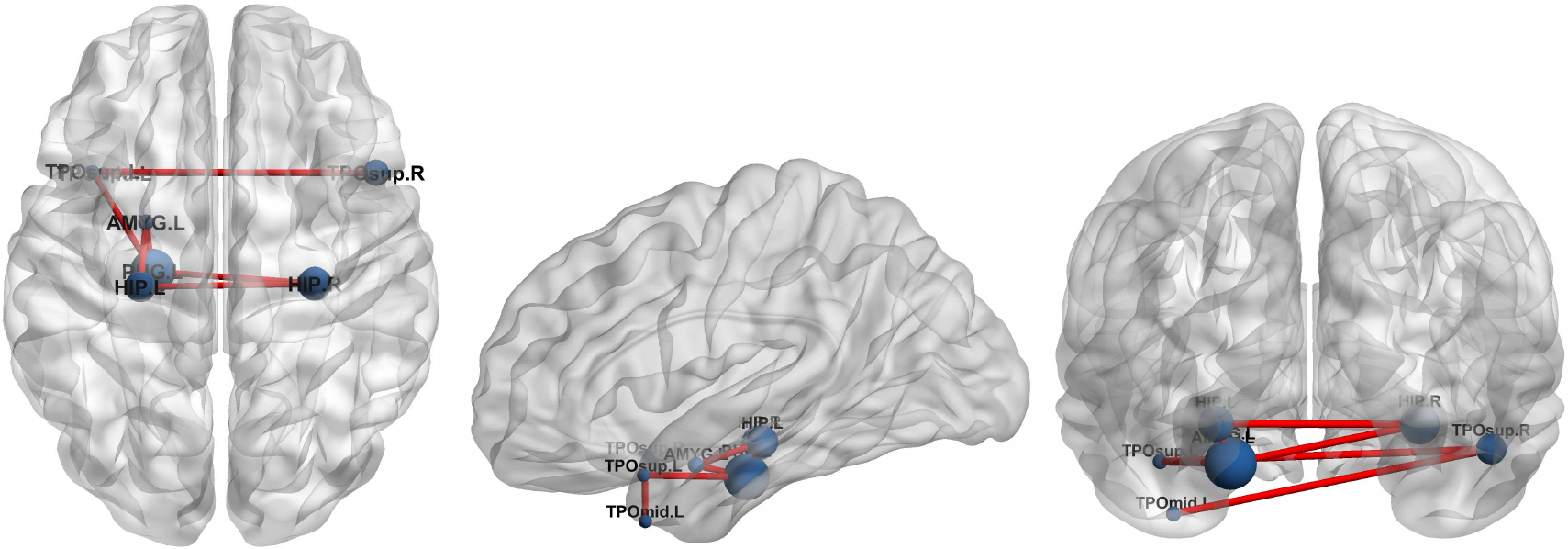}
\caption{The NC subject whose subject ID is 014\_S\_6148.} \label{fig3}
\end{figure}

\begin{figure}
\centering
\includegraphics[width=0.85\textwidth]{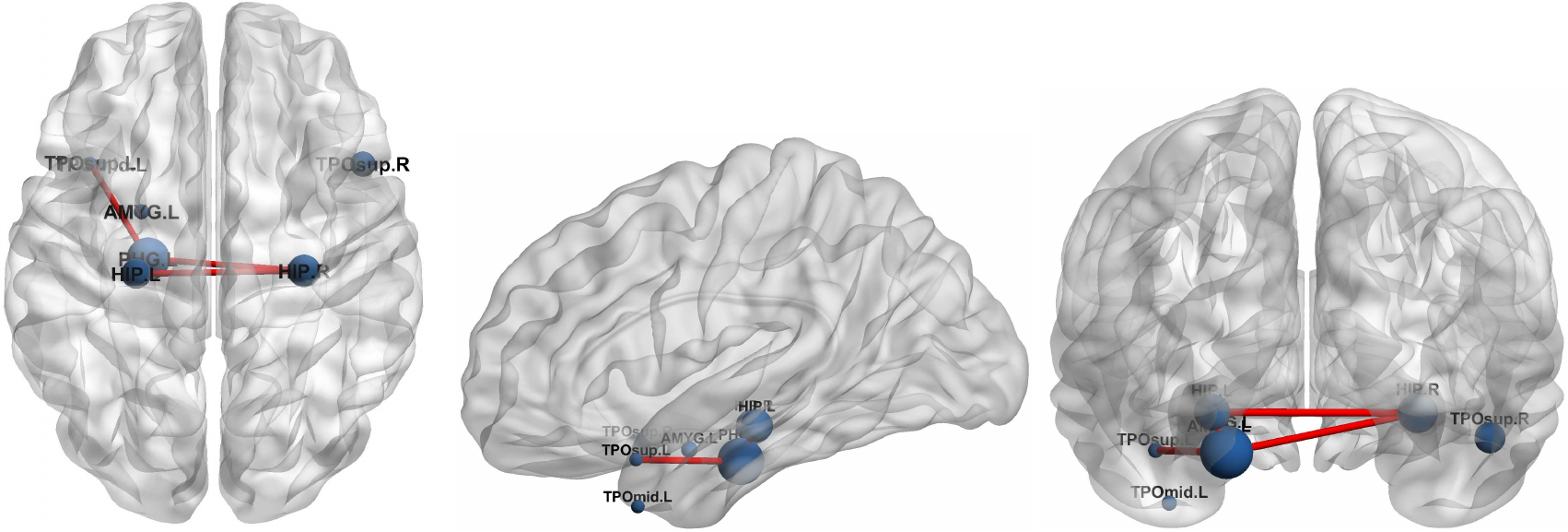}
\caption{The AD subject whose subject ID is 037\_S\_6216.} \label{fig4}
\end{figure}

We compare the average correlation coefficient in eq.\eqref{2} between each corresponding node of the AD group and the NC group. The result are shown in Fig~\ref{fig2}. We choose the top 7 nodes in Fig~\ref{fig2} (c), the IDs of these node are 37,38,39,41,83,84, and 87, which represent Hippocampus\_L, Hippocampus\_R, ParaHippocampal\_L, Amygdala\_L, Temporal\_Pole\_Sup\_L, Temporal\_Pole\_Sup\_R, and Temporal\_Pole\_ Mid\_L. We can see that these brain regions are mainly concentrated on the memory and reasoning areas, which are highly related to the AD according to the clinical studies~\cite{res}. Fig~\ref{fig3} and Fig~\ref{fig4} show the connection relationship of these top 7 brain regions by using MC matrix. We display the connection relationship graph of brain regions from the perspective of the sagittal plane view, axial plane view, and coronal plane view.

\section{Conclution}
In this paper, we proposed a novel hypergraph generative adversarial network (HGGAN) to generate individual multimodal connectivity matrix by using corresponding rs-fMRI and DTI data. We designed the Interactive Hyperedge Neurons module such that the generators can efficiently capture the complex relationship between rs-fMRI and DTI. Moreover, we proposed the Optimal Hypergraph Homomorphism algorithm to construct hypergraph structure data, which significantly improved the robustness of the generation results. The analyses of the experimental results proved that the proposed method successfully extracted the interrelated hidden structures and complementary information from different modal data. Although this paper only focuses on AD,  it is worth mention that the proposed model can be easily extended to other neurodegenerative disease.

\subsubsection*{Acknowledgement.}
This work was supported by the National Natural Science Foundations of China under Grant 61872351, the International Science and Technology Cooperation Projects of Guangdong under Grant 2019A050510030, the Distinguished Young Scholars Fund of Guangdong  under Grant 2021B1515020019, the Excellent Young Scholars of Shenzhen under Grant RCYX20200714114641211 and  Shenzhen Key Basic Research Project under Grant JCYJ20200109115641762.

\end{document}